\documentclass[runningheads]{llncs}

\usepackage[english]{babel}
\usepackage[utf8]{inputenc}
\usepackage{cite}
\usepackage{amsmath,amssymb,amsfonts}

\usepackage{algpseudocode,algorithm}
\usepackage{graphicx}
\usepackage{mathptmx} 
\usepackage{multirow}
\usepackage{subfig}
\usepackage{textcomp}
\usepackage{xcolor}
\usepackage{orcidlink}
\usepackage{wrapfig}
\usepackage{booktabs}
\def\BibTeX{{\rm B\kern-.05em{\sc i\kern-.025em b}\kern-.08em
T\kern-.1667em\lower.7ex\hbox{E}\kern-.125emX}}
\hypersetup{
    colorlinks,
    linkcolor={red!50!black},
    citecolor={blue!50!black},
    urlcolor={blue!80!black}
}

\begin{document}

\title{Quality over Quantity: Semi-Supervised Detection of Illicit Bitcoin Flows via Feature Engineering}

\author{Yekaterina Smolenkova\inst{1} \and
Nickolay Larionov\inst{2} \and
Nikolay Ivanov\inst{1} \and
Yury Yanovich\inst{1}}
\titlerunning{Semi-Supervised Detection of Illicit Bitcoin Flows}
\authorrunning{E. Smolenkova et al.}
%

\institute{Skolkovo Institute of Science and Technology, Moscow, Russia \and
Moscow Institute of Physics and Technology, Moscow, Russia}

\maketitle

\begin{abstract}

    Detecting illicit cryptocurrency transactions is hampered by extreme class imbalance, adversarial obfuscation, and a scarcity of reliable labels. While semi-supervised learning (SSL) offers a promising solution by leveraging unlabeled data, we show that its success is not guaranteed by data volume alone but is contingent on data quality. We introduce an SSL framework for detecting illicit Bitcoin flows in Shared Send Mixers (SSM) transactions, built on a comprehensive historical dataset comprising 163 million transactions. Our main conclusion is that the success of SSL depends on data quality rather than volume: high-fidelity features such as KeyLinker address clustering and Shared Send Untangling (SSU) complexity metrics achieve an F1 score of 0.84 on unlabeled data. Finally, we empirically show that common heuristics like One-Time Change (OTC), though abundant, introduce noise, while strategic reliance on higher-fidelity features like KeyLinker is essential. Our work establishes that in blockchain forensics, the path to better performance lies in smarter feature engineering for data quality, not just larger datasets.

    \keywords{Bitcoin \and Blockchain \and Shared Send Mixer \and Semi-supervised learning}
\end{abstract}

\section{Introduction}

Bitcoin's decentralized architecture provides users with pseudonymity through cryptographic addresses, enabling financial autonomy without intermediaries, but this has made it a vehicle for illicit activities, including money laundering, terrorist financing, and darknet markets. In 2024, addresses associated with illegal cryptocurrency activity generated \$40 billion~\cite{Chainalysis2025}, but this figure does not fully reflect the complexity of the investigative challenge: the transactions requiring the most urgent investigation (those passing through mixing services) are precisely the ones for which reliable factual information is most scarce. 


The Unspent Transaction Output (UTXO) model forms Bitcoin's transactional backbone \cite{Nakamoto2008, Delgado-Segura2019, Lipton2021}, where each transaction consumes existing outputs and creates new ones. Like physical banknotes, users must provide inputs covering both payment amount and miner fees, enabling privacy techniques while complicating tracing efforts. This model enables privacy-enhancing techniques like CoinJoin \cite{Maxwell2013} while simultaneously complicating transaction tracing.

CoinJoin, a prominent transaction-mixing protocol introduced in 2013, exemplifies the dual-use challenge of privacy technologies. By aggregating multiple payments into a single transaction, it severs observable links between senders and receivers through input-output obfuscation. While serving legitimate privacy needs, this Shared Send Mixer (SSM) technique is weaponized by criminals to conceal illicit fund flows from wash trading, darknet markets, and ransomware operations \cite{IOCTA2020, IOCTA2021}. The computational hardness of untangling these transactions \cite{Atlas2014, Yanovich2016b} creates analytical blind spots for law enforcement.

Existing detection methodologies show promise yet face fundamental limitations. While graph neural networks~(GNNs) and ensemble methods achieve over 90\% accuracy in conventional flows, these supervised approaches require extensive labeled datasets - a critical barrier for analyzing mixed transactions due to CoinJoin's inherent complexity and the scarcity of reliable ground truth. This limitation creates a fundamental impasse: the very transactions requiring the most scrutiny (mixed flows) have the least available labeled data. Semi-supervised learning presents a compelling alternative by leveraging both limited labeled data and abundant unlabeled records, as demonstrated in financial fraud detection \cite{SunYin2017} and network anomaly analysis~\cite{Zhang2020}.


This study advances CoinJoin transaction forensics by reframing SSL as a \textbf{data-quality problem} rather than a data-quantity problem. We address this gap through three interlocking components: a complete historical dataset of CoinJoin transactions spanning Bitcoin's full history to block 882,421; novel structural features derived from cryptographic key reuse (KeyLinker)\cite{keylinker} and transaction untangling complexity (SSU)\cite{Larionov2024}; and a confidence-calibrated 
pseudo-labeling framework that accepts new training examples only at precision-verified thresholds ($\tau^{+} \geq 0.95$), expanding coverage without propagating label noise. Together, these establish a central empirical claim: in blockchain forensics, \emph{feature quality} is the binding constraint on SSL performance, not dataset size.

The remainder of this paper is structured as follows: Section~\ref{sec:background} examines Bitcoin's UTXO transaction model and key anonymization techniques. Section~\ref{sec:related_work} analyzes existing blockchain forensic approaches and CoinJoin detection challenges. Section~\ref{sec:problem} formally defines the illicit transaction identification problem and evaluation framework. Section~\ref{sec:methodology} details our three-phase approach combining transaction clustering, feature engineering, and semi-supervised learning. Section~\ref{sec:experiments} presents comparative results across multiple detection paradigms. Section~\ref{sec:discussion}  discusses the implications of our findings and limitations. Section~\ref{sec:conclusion} concludes.

\section{Background: Bitcoin anonymization techniques}
\label{sec:background}
\subsection{Transaction model}
Bitcoin operates under a UTXO (Unspent Transaction Output) model, where each transaction consumes previous outputs as inputs and produces new outputs. Each output is associated with a script defining the conditions for spending. This design facilitates transaction chaining and allows for flexible ownership and payment schemes. However, the visibility of all transactions on the public blockchain also means that the flow of funds can be observed and analyzed.


\begin{figure}[h!]
\centering
\includegraphics[width=0.8\textwidth]{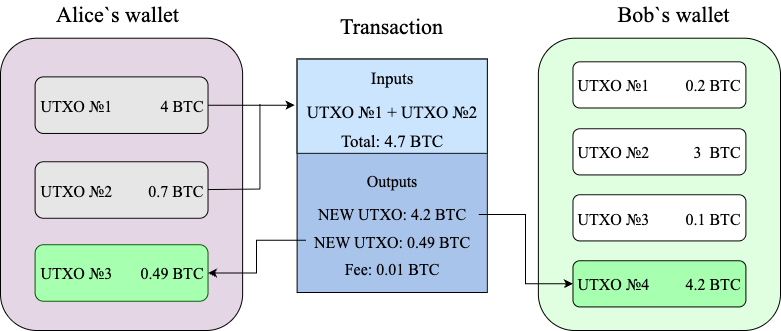}
\caption{Bitcoin UTXO transaction model: each transaction consumes previous outputs and creates new ones.}
\label{fig:tx_model}
\end{figure}

\subsection{Address clustering}
Despite the pseudonymous nature of Bitcoin addresses, several heuristic techniques can group addresses likely controlled by the same entity. Two of the most established methods include:

\begin{itemize}
\item \textbf{Common Spending (CS) heuristic:} 
If multiple addresses appear as inputs in the same transaction with a single output, they are likely controlled by the same user, since spending requires the corresponding private keys.
\item \textbf{One-Time Change (OTC) Heuristic:} In typical payment transactions, one output goes to the recipient and the other returns change to the sender. When this change address is used only once, it can help identify wallet behavior.
\end{itemize}

These heuristics underpin most clustering techniques and have been validated in academic literature and blockchain analytics platforms.

The \textbf{KeyLinker} address clustering method groups addresses by identifying the reuse of public keys across various Bitcoin address formats (P2PKH, P2SH, P2WPKH, P2WSH, P2TR). If the same public key appears in any transaction input, all addresses that can be derived from it are linked together.
Unlike the heuristic CS and OTC methods, KeyLinker inherently produces no false merges, since the identity of public keys is cryptographically proven.

\begin{figure}[h!]
\centering
\includegraphics[width=0.8\textwidth]{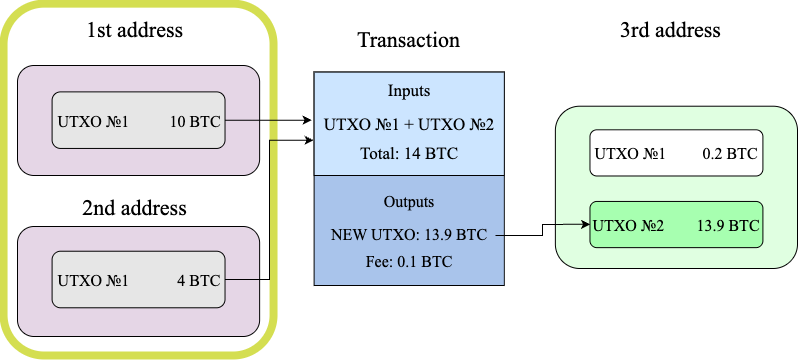}
\caption{Example of address clustering using CS heuristic. 1st address and 2nd address are owned by a single wallet}
\label{fig:clustering-cs}
\end{figure}

\begin{figure}[h!]
\centering
\includegraphics[width=0.8\textwidth]{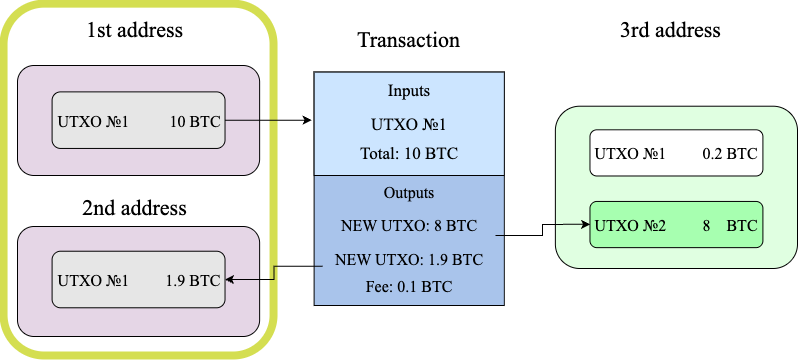}
\caption{Example of address clustering using OTC heuristic.}
\label{fig:clustering-otc}
\end{figure}

\subsection{Shared Send mixer transactions}
Shared Send refers to a class of anonymization techniques based on the CoinJoin concept. In CoinJoin, multiple users collaboratively create a single transaction where inputs and outputs are pooled together. This makes it difficult to determine which output belongs to which input, thus obfuscating the flow of funds.

A Shared Send transaction typically features many inputs and multiple outputs of the same denomination. These transactions are often constructed using special-purpose wallets or services (e.g., Wasabi Wallet) designed to facilitate anonymity.

Such transactions appear organically on the blockchain due to growing user adoption of privacy tools. However, they can also be used by illicit actors to obfuscate traces of illicit activity, such as darknet market payments or ransomware.

Despite their goal of anonymity, Shared Send transactions are susceptible to partial deanonymization. 


To analyze Shared Send transactions systematically, we adopt the \textbf{Shared Send Untangling (SSU)} framework. Each transaction is assigned to one of the following categories based on its structural complexity:
\begin{itemize}
\item \textbf{Regular:} Has either fewer than two inputs or outputs after grouping by addresses, thus nothing to untangle.
\item \textbf{Simple:} One-to-one mapping from input to output is uniquely identifiable.
\item \textbf{Separable:} Transaction decomposes into disjoint sender-receiver subgroups.
\item \textbf{Ambiguous:} Multiple plausible mappings exist between inputs and outputs.
\item \textbf{Time-Limited:} The transaction is too computationally expensive to untangle reliably.
\end{itemize}

Understanding these patterns is crucial for robust detection of anonymization schemes and building resilient forensic models.

\section{Related work}
\label{sec:related_work}

Bitcoin was originally conceived as a system with enhanced privacy through pseudonymous addresses and the absence of intermediaries~\cite{Nakamoto2008}. However, the public nature of the blockchain makes it possible to analyze transactions on a large scale, and early work has shown that user privacy is vulnerable: addresses can be partially deanonymized through analysis of behavior and transaction graph topology~\cite{Androulaki2013, Vallarano2020}. This contradiction between pseudonymity and transparency has motivated extensive research on detecting illicit transactions/addresses using supervised and semi-supervised learning.

A key class of anonymizing mechanisms is CoinJoin~\cite{Maxwell2013}, which mixes the inputs and outputs of multiple participants in a single transaction, making it difficult to trace the flow of funds. On the other hand, CoinJoin is also actively used by criminals to hide the origin of funds. Research on CoinJoin and Shared Send emphasizes that the key difficulty here is the structural ambiguity of input-output matching and the computational complexity of untangling mixed transactions~\cite{Yanovich2016b, Larionov2023, Larionov2024}. These works show that the forensic analyzability of CoinJoin strongly depends on the complexity class of the transaction. 

In parallel, address clustering developed, linking pseudonymous addresses to entities and services. Approaches evolved from simple heuristics to more advanced ML methods~\cite{Ermilov2017, Moser2022, Liu2023}. On the illicit activity detection side, modern methods (including semi-supervised learning and ensembles) demonstrate high quality on large-scale datasets: for example, semi-supervised graph models show strong results (92\% accuracy) for binary classification~\cite{Nerurkar2022} and gradient ensembles for multi-class entity categorization (91\% accuracy)~\cite{Nerurkar2021}. 

Classic ML models and activity/UTXO age features were also used for mixers~\cite{Rathore2022}. Comparative studies confirm the effectiveness of ensembles (RF, trees, SVM) in detecting suspicious transactions~\cite{Alarab2020} and report consistent results up to 87\% accuracy in various settings~\cite{Lin2022}. Crucially, existing clustering methods either introduce false merges (CS, OTC) or have not been integrated as ML features - a gap our KeyLinker-based feature engineering directly addresses. 

Neural network models also show potential in finding hidden patterns~\cite{SunYin2017, Nan2018}, including more complex graph architectures that account for heterogeneous types of connections and improve detection quality~\cite{Song2023, Lee2024}. Finally, clustering criminal communities within the transaction graph helps identify stable groups and key nodes interacting with mixing transactions~\cite{Wahrstatter2023b}. A separate line of work is devoted to improving quality through feature selection, including quantum-inspired methods~\cite{MingFong2024}. However, none of these approaches address the label scarcity specific to CoinJoin transactions, where the mixing structure itself prevents reliable ground-truth assignment.

CoinJoin remains a particularly challenging case due to structural ambiguity and label scarcity~\cite{Yanovich2016b, Larionov2023} - a gap this work directly addresses by prioritizing feature quality over data volume.


\section{Problem statement}
\label{sec:problem}

We formulate the illicit transaction detection task as a binary classification problem over Bitcoin transactions. Let $\mathcal{T}$ denote the universe of all Bitcoin transactions. Our goal is to learn a classifier:
\[
f: \mathcal{T} \rightarrow \{0,1\}
\]
where $f(t) = 1$ indicates that transaction $t \in \mathcal{T}$ is illicit (e.g., associated with mixing services, darknet markets, or scams), and $f(t) = 0$ otherwise.

Each transaction $t \in \mathcal{T}$ is represented through its native UTXO structure:
\begin{itemize}
    \item $\mathcal{I}_t = \{(a_n, A_n)\}_{n=1}^N$: Input UTXO multiset, where $a_n \in \mathbb{R}_{\geq 0}$ is the scalar input amount and $A_n \in \mathcal{A}$ is the source address
    \item $\mathcal{O}_t = \{(b_m, B_m)\}_{m=1}^M$: Output UTXO multiset, where $b_m \in \mathbb{R}_{\geq 0}$ is the output amount and $B_m \in \mathcal{A}$ is the destination address
\end{itemize}

Addresses carry semantic tags from external sources and clustering heuristics:
\[
\operatorname{Tag}: \mathcal{A} \to \left(\mathcal{L} \cup \{\bot\}\right) \times \left(\mathcal{C} \cup \{\bot\}\right),
\]
where
\begin{itemize}
    \item $\mathcal{L} = \{\text{exchange}, \text{mixer}, \text{darknet}, \text{gambling}, \ldots\}$ are entity labels
    \item $\mathcal{C} = \{\text{illicit}, \text{licit}\}$ are legitimacy labels
    \item $\bot$ indicates missing labels.
\end{itemize}
Tags propagate through clustering relationships ($\sim$): 
\[
\forall A, A' \in \mathcal{A}: A \sim A' \implies \operatorname{Tag}(A) = \operatorname{Tag}(A')
\]
The clustering relationship is established by KeyLinker public key associations, CS and OTC heuristics. 

Bitcoin transactions may contain repeated addresses in their inputs and outputs - a potentially useful characteristic for classification. We preserve this raw UTXO structure while enabling complexity analysis through strategic simplification \cite{Yanovich2016b}:
$
t \mapsto t_{\text{sim}} = \texttt{Simplify}(t).
$
This mapping groups UTXOs by addresses and their clustering relationships exclusively to determine the transaction's untangling class $\kappa(t) \in$ $\{\text{regular},$ $\text{simple},$ $\text{separable},$ $\text{ambiguous},$ $\text{time-limit}\}$ and untangling-related features. 
The $\kappa(t)$ classification feeds into the feature engineering pipeline as critical SSU attributes, while the original address repetitions remain preserved in $I_t$ and $O_t$ for feature extraction.

\section{Methodology}
\label{sec:methodology}

Our methodology is designed to identify illicit CoinJoin transactions in the Bitcoin blockchain by leveraging both supervised and semi-supervised learning techniques, enhanced by heuristic clustering and extensive feature engineering.

\subsection{Data Collection, Labeling and Feature Engineering}
The dataset includes Bitcoin blockchain data collected from the Bitcoin Core up to block 882,421 (dated February 6, 2025). 
We treat a transaction as Shared Send if it has more than one \emph{unique} input address and more than one \emph{unique} output address after address deduplication:
\[
|\mathrm{uniq}(\{A_n\}_{n=1}^N)| > 1 \quad \text{and} \quad |\mathrm{uniq}(\{B_m\}_{m=1}^M)| > 1.
\]
All CoinJoin counts in this paper refer to this definition.

To enhance our analysis, we integrated address labels from services including WalletExplorer~\cite{walletexplorer}, Elliptic++~\cite{ellipticpp}, MBAL~\cite{mbal}, and Kaggle datasets~\cite{kaggle_mbal,kaggle_labeled}, categorizing addresses by service type (exchanges, mixers, gambling, services, and mining pools) and their legality.

Our dataset comprises approximately 1.15 billion transactions, out of which 163 million are CoinJoin transactions, with 4.6 million explicitly labeled (Table~\ref{tab:comprehensive_stats}). The dataset contains 1.37 billion unique Bitcoin addresses, including 33,229 illicit and 251,083 licit addresses (Table~\ref{tab:comprehensive_stats}). The class distribution reflects a fundamental challenge: illicit CoinJoin transactions constitute approximately 12\% of labeled data, with the remainder licit. This imbalance is not an artifact of sampling but reflects the actual composition of mixing activity on the Bitcoin network, where privacy-seeking legitimate users substantially outnumber illicit actors.

\begin{table}[t]
    \centering
    \scriptsize 
    \caption{Comprehensive dataset statistics (up to block 882,421).}
    \label{tab:comprehensive_stats}
    \begin{tabular}{l r @{\hskip 0.5cm} l r}
    \toprule
    \multicolumn{2}{c}{\textbf{Transactions}} & \multicolumn{2}{c}{\textbf{Addresses}} \\
    \midrule
    Total          & 1,150.9M              & Total         & 1,370.1M        \\
    With any label        & 161.2M         & With any label       & 39.0M           \\
    CoinJoin       & 163.4M         \\
    Labeled CJ     & 4.6M           \\
    \midrule
    \multicolumn{2}{c}{\textbf{Clusterized transactions}} & \multicolumn{2}{c}{\textbf{Clusterizes addresses}}\\
    \midrule
    OTC & 188.9M &  KeyLinker &  131.4K \\
                 & &  CS-covered  & 859.0M \\ 
                 & & OTC-covered & 472.3M \\
    \midrule
    \multicolumn{4}{l}{\textbf{SSU complexity classification (transactions)}} \\
    \midrule
    Simple     & 99.1M & Separable  & 24.2M \\
    Ambiguous  & 10.5M & Time-limit  & 5.4M \\
    Regular    & 24.3M &                     &      \\
    \midrule
    \multicolumn{2}{c}{\textbf{Legality labels (addresses) }} & \multicolumn{2}{c}{\textbf{Service categories (addresses)}} \\
    \midrule
    Illicit        & 33.2K          & Service        & 18.2M \\
    Licit          & 251.1K         & Exchange       & 114.7M \\
                   &                & Gambling       & 13.2M \\
                   &                & Mixer          & 11.5M \\
                   &                & Mining         & 1.1M  \\
    \bottomrule
    \end{tabular}
\end{table}

We manually resolved duplicates and conflicting labels, addressing ambiguities such as addresses tagged simultaneously as mixers and exchanges. Addresses were clustered using three methods applied in series:
\textbf{KeyLinker} (cryptographic key reuse, no false merges), \textbf{CS}, and \textbf{OTC} heuristics. The off-chain feature group uses cluster membership from 
all three methods; Section~\ref{sec:experiments} shows that KeyLinker and CS improve detection while OTC consistently fails to do so.

Transaction labels $f(t) \in \{\text{illicit}, \text{licit}, \bot\}$ are assigned as defined in Section~\ref{sec:problem}: a transaction inherits the \emph{illicit} tag if any participating address carries it; \emph{licit} if no illicit address is present but at least one licit address is; and $\bot$ otherwise. Transactions with $f(t) = \bot$ are treated as unlabeled and used only in the pseudo-labeling stage.

\paragraph{Feature groups.} We structured the feature set into four groups:
\begin{enumerate}
    \item \textbf{UTXO attributes}: 7 features, including average UTXO lifetime, number of inputs, and outputs. 

    \item \textbf{Transaction values}: 17 features, including the sum of inputs and outputs, fee value, presence of identical amounts (has\_ident\_value), and market volume concentration index (market\_conc).

    \item \textbf{Address-level characteristics}: 6 features, indicating that the same I/O addresses were found in transaction (has\_rep\_adr\_input/output), the maximum number of times identical addresses are found on the input (cnt\_max\_rep\_adr), and other metrics of address activity.

    \item \textbf{Extended features}: In addition to the basic groups, the following attributes have been introduced:
    \begin{itemize}
      \item \textbf{SSU}: assigns each CoinJoin transaction to a structural complexity class (\texttt{regular}, \texttt{simple}, \texttt{separable}, \texttt{ambiguous}, \texttt{time-limited}) based on untangling tractability.
      \item \textbf{Off-chain}: the number of addresses in transaction links to different services (cnt\_adr\_service), exchanges (cnt\_adr\_exchange), miners (cnt\_adr\_mining), gambling (cnt\_adr\_gambling), and mixers (cnt\_adr\_mixer)  obtained using clusters via the KeyLinker method and adding cs and otc heuristics in series.
    \end{itemize}
    
\end{enumerate}

Continuous features were normalized via StandardScaler, categorical features were represented by one-hot coding, and class imbalances were compensated using class weighting in the models.

\subsection{Classification Algorithms and Training}

We partitioned the 4.62~million labeled CoinJoin transactions into training (80\%), validation (10\%), and held-out test (10\%) sets with stratified sampling. 

Given the 12\% illicit rate, accuracy is not an appropriate metric - a trivial licit-always classifier achieves 88\% accuracy with zero forensic utility. We therefore optimize F1-score and report ROC~AUC and Precision-Recall AUC as secondary metrics. F1-score is the primary optimization target as it penalizes models that sacrifice either precision or recall, critical in the forensic context where false negatives risk undetected criminal flows and false positives overload analysts, unlike accuracy, which is dominated by the licit majority class. The strong F1 scores and balanced confusion matrices of the best ensemble models confirm their ability to navigate this trade-off effectively.

Four model families were evaluated: \textbf{XGBoost}, \textbf{CatBoost}, \textbf{Random Forest}, and \textbf{Logistic Regression}. Hyperparameters were selected via stratified 5-fold cross-validation with \texttt{class\_weight=balanced}. SMOTE and ADASYN oversampling were deliberately excluded: pseudo-labeling (below) introduces new positive examples, making prior oversampling redundant and potentially double-counting.

\subsubsection*{Confidence-Gated Pseudo-Labeling}

\begin{figure}[!hbt]
\centering
\includegraphics[width=0.7\textwidth]{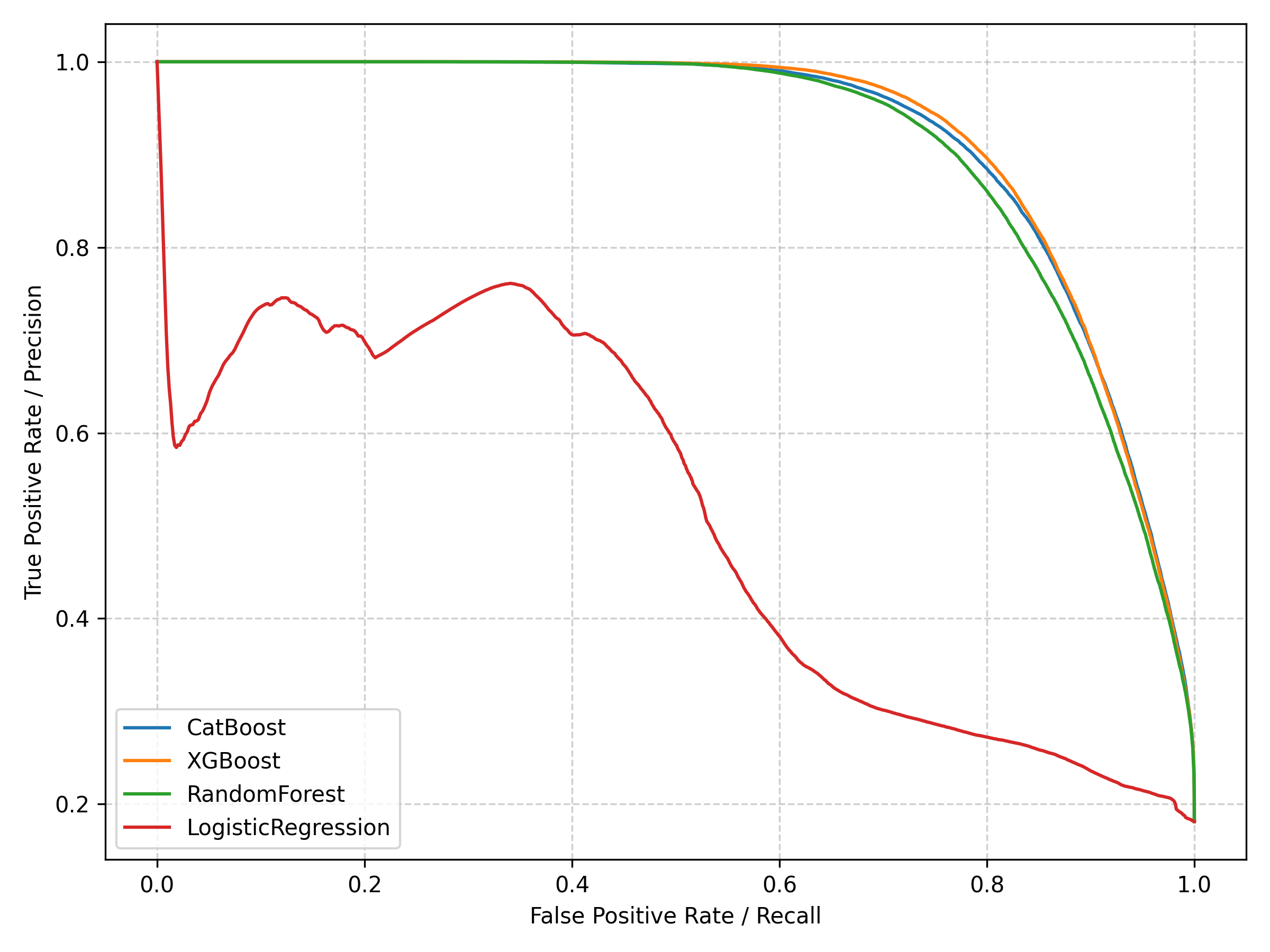}
\caption{Precision-Recall curves}
\label{fig:pr_curves}
\end{figure}

To exploit the large pool of unlabeled CoinJoin transactions, we adopt a confidence-based pseudo-labeling strategy. The trained supervised model is applied to unlabeled data, and transactions with confident scores are added to the training set. The confidence thresholds $\tau_{+}$ and $\tau_{-}$ are selected based on the Precision-Recall curve computed on the held-out validation set. This approach is preferred over fixed thresholds because the optimal operating point depends on the class imbalance of the specific feature configuration: a threshold of 0.5 would admit too many noisy pseudo-positives given the ~12\% illicit rate. By reading $\tau_{+}$ from the high-precision region of the PR curve (Figure~\ref{fig:pr_curves}), we ensure that newly added illicit examples are highly reliable. Symmetrically, $\tau_{-}$ is set at the point where recall for the licit class is maximised without admitting borderline cases. 

\begin{algorithm}[H]
\caption{Confidence-gated pseudo-labeling}
\label{alg:pseudolabel}
\begin{algorithmic}[1]
    \State \textbf{Input:} Labeled set $L$, unlabeled set $U$, 
           model $f$, thresholds $\tau^{+},\tau^{-}$
    \State Train initial model $f$ on $L$
    \ForAll{$x \in U$}
        \State $p \leftarrow f(x)$
        \If{$p \geq \tau^{+}$} add $(x,1)$ to $L$
        \ElsIf{$p \leq \tau^{-}$} add $(x,0)$ to $L$
        \EndIf
    \EndFor
    \State Retrain $f$ on updated $L$
\end{algorithmic}
\end{algorithm}

Across experiments, this procedure yielded $\tau_{+} \in [0.80, 0.98]$ and $\tau_{-} \in [0, 0.10]$ depending on model and feature set, adding hundreds of thousands to several million pseudo-labeled examples per configuration. We evaluate this methodology in Section~\ref{sec:experiments}.

\section{Numerical experiments}
\label{sec:experiments}

\subsection{Supervised Training Phase}

All experiments were conducted on a high-performance server configured with 200 GB RAM and Intel® Core™ i9-14900KF × 32 CPUs. 
We first assess the effectiveness of our feature engineering and modeling approach in a fully supervised setting. The goal at this stage is to establish how well the available labeled data can distinguish illicit from licit CoinJoin transactions, and to benchmark a set of classifiers before incorporating unlabeled examples via pseudo-labeling.

Four model types were evaluated: XGBoost, CatBoost, Random Forest and Logistic Regression. Hyperparameters were selected via stratified cross-validation (Table~\ref{tab:hyperparameters}).

\begin{table}[t]
    \centering
    \caption{Hyperparameter search spaces}
    \label{tab:hyperparameters}
    \begin{tabular}{|l|l|}
    \hline
    Model & Hyperparameters \\
    \hline
    CatBoost & learning\_rate, l2\_leaf\_reg, iterations, \\
     & depth, border\_count \\
     \hline
    XGBoost & subsample, n\_estimators, max\_depth,  \\
     & learning\_rate, gamma \\
     \hline
    RandomForest & n\_estimators, max\_depth, \\
     & min\_samples\_leaf \\
     \hline
    LogisticRegression & penalty, l1\_ratio, C \\
    \hline 
    \end{tabular}
\end{table}

\subsection{Supervised learning} 
We evaluated each model on validation and hold-out datasets. Metrics included ROC AUC, precision, recall, and F1-score values for classification (Table \ref{tab:all_models_metrics}).

\begin{table*}[t]
    \centering
    \caption{Metrics by feature set for all models}
    \label{tab:all_models_metrics}
    \scriptsize
    \setlength{\tabcolsep}{3pt}
    \begin{tabular}{l *{5}{c} cccc}
    \toprule
    \multicolumn{1}{c}{\textbf{Model}} &
    \multicolumn{5}{c}{\textbf{Features}} &
    \multicolumn{4}{c}{\textbf{Metrics}} \\
    \cmidrule(lr){1-1}\cmidrule(lr){2-6}\cmidrule(lr){7-10}
    & DEFAULT & REUSE & CS & OTC & SSU & Precision & Recall & F1-score & ROC AUC \\
    \midrule
    \multirow{7}{*}{CatBoost}
     & \checkmark &           &           &           &           & 0.929 & 0.689 & 0.791 & 0.958 \\
     & \checkmark & \checkmark &           &           &           & 0.929 & 0.730 & 0.818 & 0.966 \\
     & \checkmark & \checkmark & \checkmark &           &          & 0.930 & 0.740 & 0.824 & 0.969 \\
     & \checkmark & \checkmark & \checkmark & \checkmark &        & 0.928 & 0.740 & 0.823 & 0.967 \\
     & \checkmark &           &           &           & \checkmark & 0.926 & 0.705 & 0.800 & 0.960 \\
    & \checkmark & \checkmark & \checkmark &  & \checkmark & 0.936 & 0.746 & \textbf{0.830} & 0.970 \\
     & \checkmark & \checkmark & \checkmark & \checkmark & \checkmark & 0.930 & 0.745 & 0.827 & 0.968 \\
     
    \midrule
    \multirow{7}{*}{XGBoost}
     & \checkmark &           &           &           &           & 0.875 & 0.762 & 0.814 & 0.959 \\
     & \checkmark & \checkmark &           &           &           & 0.888 & 0.790 & 0.837 & 0.967 \\
     & \checkmark & \checkmark & \checkmark &           &          & 0.897 & 0.796 & 0.844 & 0.970 \\
     & \checkmark & \checkmark & \checkmark & \checkmark &        & 0.895 & 0.792 & 0.841 & 0.968 \\
     & \checkmark &           &           &           & \checkmark & 0.882 & 0.767 & 0.821 & 0.961 \\
      & \checkmark & \checkmark & \checkmark &  & \checkmark & 0.900 & 0.797 & \textbf{0.845} & 0.970 \\
     & \checkmark & \checkmark & \checkmark & \checkmark & \checkmark & 0.901 & 0.788 & 0.840 & 0.969 \\
     
    \midrule
    \multirow{7}{*}{RandomForest}
     & \checkmark &           &           &           &           & 0.883 & 0.739 & 0.804 & 0.957 \\
     & \checkmark & \checkmark &           &           &           & 0.906 & 0.743 & 0.816 & 0.962 \\
     & \checkmark & \checkmark & \checkmark &           &          & 0.899 & 0.769 & 0.829 & 0.967 \\
     & \checkmark & \checkmark & \checkmark & \checkmark &        & 0.908 & 0.739 & 0.815 & 0.960 \\
     & \checkmark &           &           &           & \checkmark & 0.893 & 0.731 & 0.805 & 0.957 \\
      & \checkmark & \checkmark & \checkmark &  & \checkmark & 0.901 & 0.769 & \textbf{0.830} & 0.967 \\
     & \checkmark & \checkmark & \checkmark & \checkmark & \checkmark & 0.907 & 0.744 & 0.818 & 0.962 \\
     
    \midrule
    \multirow{7}{*}{Logistic Regression}
     & \checkmark &           &           &           &           & 0.724 & 0.103 & \textbf{0.180} & 0.665 \\
     & \checkmark & \checkmark &           &           &           & 0.650 & 0.079 & 0.141 & 0.752 \\
     & \checkmark & \checkmark & \checkmark &           &          & 0.667 & 0.088 & 0.156 & 0.757 \\
     & \checkmark & \checkmark & \checkmark & \checkmark &        & 0.638 & 0.078 & 0.139 & 0.763 \\
     & \checkmark &           &           &           & \checkmark & 0.765 & 0.009 & 0.016 & 0.731 \\
      & \checkmark & \checkmark & \checkmark &  & \checkmark & 0.700 & 0.079 & 0.143 & 0.770 \\
     & \checkmark & \checkmark & \checkmark & \checkmark & \checkmark & 0.674 & 0.006 & 0.012 & 0.769 \\
    \bottomrule
    \end{tabular}
\end{table*}

Adding KeyLinker-derived reuse features improves F1 by 2–3 percentage points across all models. Common Spending features provide a further incremental gain. Notably, OTC features consistently fail to improve and occasionally degrade performance - a pattern we attribute to the adversarial evasion of behavioral heuristics by illicit actors (discussed in Section~\ref{sec:discussion}).

XGBoost achieves the best supervised performance with an F1-score of 0.845 (default+reuse+cs+ssu) and ROC-AUC~=~0.970, closely followed by CatBoost (F1 up to 0.830). Logistic Regression underperformed due to its inability to model nonlinear patterns.

\subsection{Semi-Supervised Learning with Pseudo-Labeling}

While supervised models performed robustly, the vast pool of unlabeled CoinJoin transactions presents an opportunity for further improvement. To leverage this, we employed a pseudo-labeling strategy: our best supervised models were used to assign labels to unlabeled data, but only for predictions with high estimated confidence.


Across models, the most important features include aggregated statistics of transaction inputs and outputs, including \texttt{fee}, \texttt{cnt\_input}/\texttt{cnt\_output}, \\ \texttt{min\_value\_input}/\texttt{output} and categorical indicators linked to service markers and Shared-Send structure (e.g., \texttt{ss\_separable}, \texttt{ss\_time\_limit},\texttt{cnt\_adr\_exchange}) consistently appear among the top features.

\begin{table*}[t]
    \centering
    \caption{Metrics by feature set on pseudo stage}
    \label{tab:all_models_metrics_pseudo}
    \scriptsize
    \setlength{\tabcolsep}{3pt}
    \begin{tabular}{l *{5}{c} cccc}
    \toprule
    \multicolumn{1}{c}{\textbf{Model}} &
    \multicolumn{5}{c}{\textbf{Features}} &
    \multicolumn{4}{c}{\textbf{Metrics}} \\
    \cmidrule(lr){1-1}\cmidrule(lr){2-6}\cmidrule(lr){7-10}
    & DEFAULT & REUSE & CS & OTC & SSU & Precision & Recall & F1-score & ROC AUC \\
    \midrule
    \multirow{7}{*}{CatBoost}
     & \checkmark &           &           &           &           & 0.848 & 0.759 & 0.801 & 0.956 \\
     & \checkmark & \checkmark &           &           &           & 0.873 & 0.775 & 0.821 & 0.964 \\
     & \checkmark & \checkmark & \checkmark &           &          & 0.866 & 0.795 & 0.829 & 0.966 \\
     & \checkmark & \checkmark & \checkmark & \checkmark &        & 0.866 & 0.788 & 0.825 & 0.964 \\
     & \checkmark &           &           &           & \checkmark & 0.856 & 0.764 & 0.807 & 0.958 \\
    & \checkmark & \checkmark & \checkmark &  & \checkmark & 0.868 & 0.803 & \textbf{0.834} & 0.968 \\
     & \checkmark & \checkmark & \checkmark & \checkmark & \checkmark & 0.874 & 0.788 & 0.829 & 0.966 \\
    \midrule
    \multirow{7}{*}{XGBoost}
     & \checkmark &           &           &           &           & 0.865 & 0.757 & 0.807 & 0.957 \\
     & \checkmark & \checkmark &           &           &           & 0.891 & 0.779 & 0.832 & 0.966 \\
     & \checkmark & \checkmark & \checkmark &           &          & 0.887 & 0.796 & 0.839 & 0.969 \\
     & \checkmark & \checkmark & \checkmark & \checkmark &        & 0.892 & 0.787 & 0.836 & 0.966 \\
     & \checkmark &           &           &           & \checkmark & 0.873 & 0.763 & 0.814 & 0.959 \\
      & \checkmark & \checkmark & \checkmark &  & \checkmark & 0.897 & 0.792 & \textbf{0.841} & 0.969 \\
     & \checkmark & \checkmark & \checkmark & \checkmark & \checkmark & 0.890 & 0.787 & 0.836 & 0.967 \\
    \midrule
    \multirow{7}{*}{RandomForest}
    & \checkmark &           &           &           &           & 0.853 & 0.757 & 0.802 & 0.955 \\ 
     & \checkmark & \checkmark &           &           &           & 0.875 & 0.762 & 0.814 & 0.961 \\
     & \checkmark & \checkmark & \checkmark &           &          & 0.877 & 0.781 & \textbf{0.826} & 0.965 \\
     & \checkmark & \checkmark & \checkmark & \checkmark &        & 0.870 & 0.765 & 0.814 & 0.959  \\
     & \checkmark &           &           &           & \checkmark & 0.858 & 0.751 & 0.801 & 0.955 \\
     & \checkmark & \checkmark & \checkmark &  & \checkmark & 0.882 & 0.777 & \textbf{0.826} & 0.965 \\
     & \checkmark & \checkmark & \checkmark & \checkmark & \checkmark & 0.872 & 0.768 & 0.817 & 0.960 \\
    \bottomrule
    \end{tabular}
\end{table*}

Table~\ref{tab:all_models_metrics_pseudo} demonstrates that pseudo-labeling with high-confidence thresholds consistently improves upon the supervised baseline reported in Table~\ref{tab:all_models_metrics}.

\begin{table*}[t]
    \smallskip
    \caption{Pseudo-labeling thresholds and the number of added pseudo-labeled samples for the default+reuse+cs+ssu setting.}
    \label{tab:pseudolabel_thresholds_main}
    \medskip
    \centering
    \begin{tabular}{lcccc}
        \hline
         Model & $\tau^{+}$ & Pseudo-labeled pos & $\tau^{-}$ & Pseudo-labeled neg \\
        \hline
         CatBoost &  0.904 &1,360,746 & 0.006 & 8,600,795 \\ 
         XGBoost & 0.982 & 1,364,991 & 0.022 & 8,627,624 \\ 
         RandomForest &  0.933 & 774,035 & 0.015 & 5,203,072 \\ 
        \hline
    \end{tabular}
    \label{tab:best_thresholds}
\end{table*} 

Table~\ref{tab:pseudolabel_thresholds_main} reports the score thresholds used for pseudo-labeling in the best-performing setting (default+reuse+cs+ssu) and the resulting number of added samples per model. As expected for highly imbalanced CoinJoin labeling, the negative threshold $\tau_{-}$ produces a substantially larger set of confident pseudo-negatives (5.2-8.6M) than pseudo-positives (0.77-1.36M). This reflects that most unlabeled CoinJoin transactions are predicted as licit with high confidence, while the illicit class remains rarer and requires a stricter acceptance criterion.

\section{Discussion}
\label{sec:discussion}
The experimental results provide several nuanced insights into the mechanics of illicit flow detection within Shared Send Mixer (SSM) transactions. Beyond the raw performance metrics, the ablation studies reveal a distinct hierarchy in feature utility that challenges the conventional wisdom of maximizing heuristic coverage.

The observed performance degradation when incorporating One-Time Change (OTC) heuristics suggests that signal noise outweighs coverage gains in adversarial settings. While OTC increases the volume of clustered addresses, it relies on behavioral assumptions that illicit actors actively evade or that simply do not hold in complex mixing scenarios. In contrast, structural features derived from Shared Send Untangling (SSU) and cryptographic public key reuse (KeyLinker) provide more robust signals because they are grounded in the immutable constraints of the UTXO model and cryptographic proofs rather than behavioral hypotheses. This indicates that for forensic models, feature engineering should prioritize resistance to adversarial manipulation over sheer volume of data points.

Furthermore, the consistent superiority of tree-based ensemble methods over linear baselines underscores the nonlinear nature of illicit transaction patterns. Illicit flows are not defined by single anomalous metrics but by complex interactions between transaction structure, value distribution, and address history. Linear models fail to capture these high-order interactions, whereas gradient boosting frameworks effectively model the decision boundaries where illicit behavior conceals itself within legitimate-looking traffic. This finding validates the computational cost of ensemble methods as necessary for achieving high recall in high-stakes forensic environments.

The success of pseudo-labeling depends critically on the confidence thresholds applied during the augmentation phase. Our results show that aggressive pseudo-labeling without strict confidence filtering propagates errors from the initial supervised model, particularly when noisy features are present. However, when guided by high-fidelity features, the pseudo-labeling process acts as a quality filter, expanding the training distribution without compromising precision. This suggests that semi-supervised learning in blockchain forensics should be viewed as a mechanism for refining signal density rather than simply inflating dataset size.

\paragraph{Limitations.} Future studies should take into account a number of limitations. First, our ground truth relies on off-chain labeling sources, which may contain inaccuracies or lag behind emerging illicit schemes. While we mitigated this by integrating multiple sources, systematic labeling noise remains a potential confounder. Second, the cryptocurrency landscape evolves rapidly; new mixing protocols beyond CoinJoin may exhibit different structural properties that require adaptive feature engineering. Finally, our evaluation focuses on historical data; real-time deployment would require optimizing the computational overhead of SSU classification for streaming transaction analysis.

\section{Conclusion}
\label{sec:conclusion}
Our central finding is that semi-supervised learning for illicit Bitcoin flow detection is a data-quality problem, not a data-quantity problem. High-fidelity structural signals, KeyLinker address clustering, and SSU complexity features consistently drove performance gains across all evaluated models, while the abundant but noisy OTC heuristic degraded results. This hierarchy held through both the supervised and pseudo-labeling phases, where confidence-gated augmentation improved recall without sacrificing precision.

In the supervised setting, XGBoost achieved the strongest performance (F1\,=\,0.845, ROC-AUC\,=\,0.970), with CatBoost and RandomForest following closely. The semi-supervised extension further validated this approach: pseudo-labeling successfully increased coverage only when constrained by high-confidence thresholds and guided by stable feature sets.

These findings have direct implications for blockchain forensic operations. As privacy-enhancing technologies become more sophisticated, reliance on simple heuristic clustering will become increasingly insufficient. Instead, forensic systems must adopt hybrid approaches that combine structural analysis, cryptographic evidence, and semi-supervised learning to maintain detection efficacy. Future work will focus on extending these feature sets to emerging privacy protocols and optimizing the framework for real-time transaction monitoring. Concretely, extending SSU complexity analysis to Whirlpool and PayJoin protocols, and validating the framework under streaming transaction conditions, represent the most pressing next steps toward operational deployment.

\bibliographystyle{splncs04}
\bibliography{bib}

\end{document}